%% file: tmlr.tex
\documentclass[10pt]{article} % For LaTeX2e
\usepackage{arxiv}
\usepackage{hyperref}
\usepackage{url}
\input{myincludes}

\title{\center{Can We Count on LLMs? The Fixed-Effect Fallacy and Claims of GPT-4 Capabilities}}

\author{Thomas Ball, Shuo Chen and Cormac Herley\\Microsoft Research\\Redmond, WA}

\def\month{09}  % Insert correct month for camera-ready version
\def\year{2024} % Insert correct year for camera-ready version
\def\openreview{\url{https://openreview.net/forum?id=qt4d0EGZsK}} % Insert correct link to OpenReview for camera-ready version
 
\usepackage{microtype}
\usepackage{graphicx}
\usepackage{subfigure}
\usepackage{booktabs} % for professional tables

\usepackage{amsmath}
\usepackage{amssymb}
\usepackage{mathtools}
\usepackage{amsthm}
\usepackage{natbib}

\usepackage[capitalize,noabbrev]{cleveref}

\theoremstyle{plain}

\theoremstyle{definition}

\theoremstyle{remark}

\usepackage[disable,textsize=tiny]{todonotes}

\begin{document}

\maketitle
\begin{abstract} 
In this paper we explore evaluation of LLM capabilities.
We present measurements of GPT-4 performance on several deterministic tasks; each task involves a basic calculation and takes as input parameter some element drawn from
a large well-defined population (e.g., count elements in a list, multiply two k-digit numbers, etc).
We examine several conditions per-task and perform enough
trials so that statistically significant differences can be detected.
This allows us to investigate the sensitivity of task-accuracy both to query phrasing and input parameter
population. 
We find that
seemingly trivial modifications in the task-prompt or input population can yield
differences far larger than can be explained by sampling
effects. For example, performance on a simple list-counting
task varies with query-phrasing and list-length, but also with list composition (i.e., the thing-to-be-counted)
and object frequency (e.g., success when an element accounts for $\approx$ 50\% of a list is different from when it
accounts for $\approx$ 70\% etc).

We conclude that efforts to quantify LLM capabilities 
easily succumb to the language-as-fixed-effect fallacy,
where experimental observations are
improperly generalized beyond what the data supports. A
consequence appears to be that intuitions that have been
formed based on interactions with humans form a 
very unreliable guide as to which input modifications 
should ``make no difference'' to LLM performance.

\end{abstract}

\input{listQueries/intro}
\input{listQueries/fixedEffects}

\input{listQueries/queries}

\input{listQueries/related}

\input{listQueries/discuss}

\appendix
\input{listQueries/appendix}

%%\bibliography{\jobname}
\bibliography{llms}

\bibliographystyle{tmlr}
%\bibliographystyle{icml2024}

%%%%%%%%%%%%%%%%%%%%%%%%%%%%%%%%%%%%%%%%%%%%%%%%%%%%%%%%%%%%%%%%%%%%%%%%%%%%%%%
%%%%%%%%%%%%%%%%%%%%%%%%%%%%%%%%%%%%%%%%%%%%%%%%%%%%%%%%%%%%%%%%%%%%%%%%%%%%%%%
% APPENDIX
%%%%%%%%%%%%%%%%%%%%%%%%%%%%%%%%%%%%%%%%%%%%%%%%%%%%%%%%%%%%%%%%%%%%%%%%%%%%%%%
%%%%%%%%%%%%%%%%%%%%%%%%%%%%%%%%%%%%%%%%%%%%%%%%%%%%%%%%%%%%%%%%%%%%%%%%%%%%%%%
\newpage
\appendix
\onecolumn

\end{document}

%% file: myIncludes.tex
\usepackage{tikz}
\usepackage{amsmath}
\usepackage{mathtools}
\usepackage{multicol}
\usepackage[framemethod=TikZ]{mdframed} 
\usepackage{amssymb}
\usepackage{bm}
\usepackage{array}

\usepackage{stfloats}
\usepackage{fixltx2e}
\usepackage{multicol}
\usepackage{multirow}
\usepackage{url}
\usepackage{verbatim}
\usepackage{wasysym}
\usepackage{rotating}
\usepackage{ifthen}
\usepackage{algorithm}
\usepackage{algorithmic}
\usepackage{balance}
\usepackage{makecell}

\usepackage{filecontents}

\newcommand{\btr}{better-than-random}  
\newcommand{\chatPrompt}[3]{\begin{mdframed}[backgroundcolor=yellow,roundcorner=3pt,linewidth=0.5mm] {\bf Prompt:} #1\\ {\bf Correct Answer:} {#2}\\ {\bf GPT-4 Answer:} {#3}\end{mdframed}} 

%% file: listQueries/intro.tex
\section{Introduction}
\label{sec:intro}

%\begin{quote}
%{\it
%The original question, `Can machines think?' I believe to be too meaningless to deserve discussion.}
%$~~~~~~~~~~~~~~~~~~~~~~~~~~~~~$ -- A. Turing \citep{Turing:1950}
%\end{quote} 
%
% 

Rapid improvements in the performance of large language models (LLMs)
have spurred great interest in evaluating their capabilities. In addition to answering general knowledge questions 
and summarizing text, GPT-4 has demonstrated the capability to compose poetry, solve chess puzzles and Geometry problems,  and perform basic coding tasks.
Capabilities that seem beyond the simple next-token-prediction they were trained on, causes some to 
suggest this as evidence of emergent behaviors from LLMs, or even 
that we may be witnessing the early signs of Artificial General Intelligence (AGI) \citep{bubeck2023sparks}. Others 
are  not convinced, and suggest that LLMs simply parrot pastiches of text snippets from their training sets \citep{bender2021dangers}.

The documentation of surprising capabilities has been accompanied by many accounts of failures. 
Hallucinations (where LLMs offer plausible but entirely invented detail) have proved hard to eliminate. Arkoudas points out that
GPT-4 struggles with some basic tasks that humans find easy or trivial; e.g., they aren't reliable even on tasks such as counting, multiplication, etc \citep{arkoudas2023gpt4}.
McCoy et al suggest that many of the remarkable capabilities are simply artifacts of the training set and autoregressive task that GPT-4 was trained to solve \citep{mccoy2023embers}.

An accumulation of observed successes and failures at particular tasks unfortunately does little to settle questions about 
LLM reliability or capabilities. %We lack the ability to predict performance at individual tasks making it difficult to evaluate reliability.
In this paper we present results on a series of deterministic tasks; each of the tasks involves a basic calculation and takes as input parameter some element drawn from a large well-defined population (e.g., count elements in a list, multiply two $k$-digit numbers, etc). Since, by construction, the correct answer is easy to determine, we can measure performance without costly and  subjective hand-labelling or assessments. By randomly sampling the input parameter populations we can measure performance on large numbers of  that are semantically and logically equivalent. Since the parameter spaces can be arbitrarily large the concern about verbatim contamination of training data is greatly reduced.
This allows us to investigate the sensitivity of task-accuracy
both to  query phrasing and input parameter population; we do this 
at sufficient scale to detect statistically significant differences.
We investigate both re-wordings of the prompt and changes to the input population.
 For example, our population might be length-$21$ lists of floating point numbers, and the task might be to find the median, but modifications might be to 
try reworded versions of the prompt, or try lists with a different number of significant decimal places given. 
% We use a default of $500$ trials per condition tested, so that even a $5$\%
% difference is generally statistically significant. 

Our contributions are as follows. We present measurements of GPT-4 performance on several deterministic tasks. We examine several 
conditions per-task and perform enough trials ($500$ per condition unless otherwise stated) so that statistically significant
differences can be detected. For all tasks and all conditions this entails about $37$k responses from GPT-4; all prompts, responses and associated metadata are openly available to those who wish to check or build upon our findings.
We measure performance on tasks such as counting, sorting, multiplication, etc, and find that accuracy, while better-than-random, is often very poor.
We find that seemingly trivial modifications both in the prompt-phrasing and parameter population can yield differences far larger than can be explained by sampling effects. For example, performance on a simple list-counting task varies with query-phrasing and list-length, but also with list composition (i.e., the thing-to-be-counted) and object frequency (e.g., success when an element accounts for $\approx 50$\% of a list is different from when it accounts for $\approx 70$\% etc).

We conclude that efforts to quantify LLM capabilities easily succumb to the language-as-fixed-effect fallacy 
\citep{clark1973language,coleman1964generalizing,yarkoni2019generalizability}, where experimental observations on language-tasks are improperly generalized beyond what the data supports. A consequence appears to be that intuitions that have been formed based on interactions with humans form a very unreliable guide as to  which modifications should ``make no difference'' to LLM performance. For example, the abstractions that we take for granted for humans (e.g., of separating the
task of counting from the thing-to-be-counted) do not appear to be replicated by LLMs.

Sensitivity of LLM performance to query phrasing has spawned efforts to improve accuracy by using few-shot, Chain-of-Thought and scratchpad techniques. However, efforts to quantify this sensitivity 
are nascent. Sclar et al examine the effect of phrasing on accuracy for multiple choice tasks using the LLaMA-2-13B model \citep{sclar2023quantifying}. Sun et al examine zero-shot robustness for the MMLU \citep{hendrycks2020measuring} and BIG-bench \citep{ghazal2013bigbench} datasets using several models having between 3B and 13B parameters. There are important points of difference between ours and previous work. First, we explore accuracy on atomic tasks such as counting and multiplication rather than on datasets of multiple-choice questions that may have been seen in training (e.g., there is evidence that GPT4 has seen the BIG-bench canary GUID \citep{bubeck2023sparks}).
Second, we use parameterized tasks and explore sensitivity to input parameters as well as prompt phrasing (e.g., showing that counting accuracy depends on the thing-to-be-counted). Finally, we evaluate using GPT-4; this has between one and two orders of magnitude more parameters than those used in \citep{sclar2023quantifying,sun2023evaluating}. This allows us to have confidence that the problems do not seem significantly alleviated by model scale.

We wish to be clear that our goal is not to determine whether LLMs can or cannot count, sort, or multiply, etc. First, we have other ways
of performing these tasks. Second, it is possible that prompt engineering, providing few-shot examples, the use of Chain-of-Thought reasoning, or the invocation of plug-ins might sometimes improve performance. However, our goal is not to improve the accuracy in particular settings. Rather, it is to draw attention to an 
unaddressed difficulty in establishing accuracy: evaluation of LLM capabilities seems particularly susceptible to
a major pitfall that exists when we go from particular experimental observations to general claims. That is, sensitivity to seemingly trivial modifications means that observed accuracy numbers cannot be assumed to generalize (even to entirely equivalent versions of a task). So, for example, while prompt engineering might yield accuracy improvement on a particular version of a task, we can't assume that that improvement will be observed in rephrased versions.
While we've demonstrated the problem on basic arithmetic tasks it seems unlikely to be confined to that domain. For example,  LLM performance at certain tasks might be improved by invoking a plug-in, writing code or using Chain-Of-Thought, but deciding when and how to do so is itself a task with success rate subject to the sensitivities we highlight. That is, invoking plugins doesn't solve even the basic counting task  if the decisions on when and which plugin to invoke is itself brittle and sensitive to prompt phrasing.

So, can LLMs count (or multiply, or sort etc)? Our evidence suggests that variation as we sample 
possible phrasings is too high to allow a Yes-or-No answer, and that accuracy estimates must be 
regarded as particular to the experimental setup used.
This also means that reporting observed performance or accuracy numbers on other deterministic tasks (such as standardized tests \citep{katz2023gpt,takagi2023performance,nori2023capabilities}, textbook problems, etc) is not sufficient to establish general capabilities.

%% file: listQueries/fixedEffects.tex
\section{Background: The Language-as-Fixed-Effect Fallacy}
\label{sec:fixedEffect}

The Language-as-Fixed-Effect Fallacy, as described by Clark \citep{clark1973language}, is the phenomenon where a claim
supported by statistical evidence does not generalize beyond the specifics of the experimental setup. He illustrates with
a language-task thought-experiment originally proposed by Coleman \citep{coleman1964generalizing}. Let ${\bm N}$ be the set of all English nouns, ${\bm V}$ the set of all verbs, and
let $T(.)$ be a test statistic representing how well humans perform at some task involving words (e.g., how well they can spell them, how quickly they can type them, etc).
Suppose that experimenter A wishes to test the hypothesis that people perform the task better on nouns than on verbs:
$$\mathcal{H}_A =   T({\bm N}) > T({\bm V}).$$
Suppose experimenter B wishes to test the opposite:
$$\mathcal{H}_B = T({\bm N}) < T({\bm V}).$$
Let's stipulate, by contrast, that they are both wrong, and that $T({\bm N}) = T({\bm V})$.

As a test of $\mathcal{H}_A$ the first experimenter selects  subsets ${\bm N}_A \subset {\bm N}$   and ${\bm V}_A \subset {\bm V}$ each with some fixed number of randomly selected nouns and verbs.
%10 nouns at random (e.g., ${\bm N}_A = \{$apple,  elephant, $\cdots\}$) and 10 verbs (e.g., ${\bm V}_A = \{$swim, shine $\cdots\}$).
With this choice she recruits participants and on finding that $T({\bm N}_A) > T({\bm V}_A)$, by a statistically significant amount,
she rejects the null hypothesis (that there's no difference) and concludes she has firm evidence in favor of  $\mathcal{H}_A.$
Similarly, the second experimenter selects  at random different subsets ${\bm N}_B \subset {\bm N}$   and ${\bm V}_B \subset {\bm V}$ with fixed numbers of nouns and verbs.
With this fixed choice he recruits participants and finds $T({\bm N}_B) < T({\bm V}_B)$, by a statistically significant amount, and concludes this is firm evidence for $\mathcal{H}_B.$

The problem is that while both A and B intend to generalize to the whole population ${\bm N}$ and ${\bm V}$ they have tested only on  particular subsets.
There is good evidence to believe that, with any collection of participants,  we could verify both $T({\bm N}_A) > T({\bm V}_A)$  and $T({\bm N}_B) < T({\bm V}_B)$, but neither of these 
is  enough to support either $\mathcal{H}_A$ or $\mathcal{H}_B$.  In the language
of statistical testing our experimenters have treated random effects as fixed \citep{clark1973language}.

Fixed effects are those that are considered constant across the relevant population, 
while random effects are those that vary (for an account of various other definitions see \citep{gelman2005analysis}). 
In the experiments above there are two populations involved: the populations of noun-verb collections,
and the population of human participants. When she generalized from ${\bm N}_A$, ${\bm V}_A$ to ${\bm N}$, ${\bm V}$
our first experimenter implicitly assumed that
 any other subsets ${\bm N}_C \subset {\bm N}$ and ${\bm V}_C \subset {\bm V}$ would also give the result that she observed
(i.e., $T({\bm N}_C) > T({\bm V}_C)).$
If this were true she'd be justified in thinking that  her observed  difference was powerful evidence for $\mathcal{H}_A.$ If this is not true then her experiment supports
 only the narrow uninteresting claim $T({\bm N}_A) > T({\bm V}_A).$ Effectively, she assumed that what she observed wasn't particular to ${\bm N}_A, {\bm V}_A$ but general to ${\bm N}, {\bm V}.$

In a colloquial sense fixed effects are ones where the particular choice doesn't affect the generality we wish to claim.
We expect, for example, that what an experimenter had for breakfast or what color socks she was wearing has no effect on the outcome; these are not details that have to be faithfully
reproduced to ensure replication of the original experiment. In this telling the fixed-effect fallacy is simply assuming that certain details don't matter when in fact they do. Unfortunately, there's no simple way to determine that a certain variable
has no influence on an experimental result; experiments necessarily involve many judgements  about which details  matter and which do not, and many of those judgements are subjective. One of our findings is that intuitions about which modifications might make a difference can be very flawed; that human performance 
remains constant under a certain modification is no guarantee at all that LLM performance also will. 

\section{The Fixed-Effect Fallacy and LLM Task Performance}
We wish to evaluate whether, and how well, an LLM can perform a particular task that has a single deterministic correct answer (e.g., counting, deciding to invoke a plug-in, or Retrieval-Augmented Generation etc). For the counting task one approach might be to produce a list of objects and prompt the LLM
to count the occurrences of a particular item. To make the experimental setup concrete we might specify a list length and dictionary of possible elements. For example:
\begin{verbatim}
rLen = 20
listOfItems = [`mango',`peach']
r = random.choices(listOfItems, k = rLen)
\end{verbatim}
%
%\begin{multline} 
%\mbox{\tt qLen = 20}\\
%\mbox{\tt listOfItems = [`mango',`peach']}\\
%\mbox{\tt q = random.choices(listOfItems, k = qLen)}
%\end{multline}
is a Python snippet that will return a length-20 list with the elements of {\tt listOfItems} chosen at random with replacement. When there are only two elements, as shown,
there's a population of $2^{20}$ such lists; call this population ${\bf R}.$
We might prompt the LLM with:
\begin{verbatim}
prompt = ``How many times does `mango' appear in this list: '' + str(r)
\end{verbatim}
%
%\begin{multline}\nonumber
%\mbox{\tt prompt = ``How many times does `mango'} \\ \mbox{\tt appear in this list:'' + str(q)}\end{multline}
%$$\mbox{\tt `How many times does mango appear in this}$$ $$\mbox{\tt list:' + str(q)}$$
where $r \in {\bm R}.$ By repeating this query with many different elements of ${\bm R}$ we
might try to build a picture of the LLM's performance at the task.

In this setup choice of list from ${\bm R}$ is being treated as the only random effect; i.e., the only source of variation \citep{gelman2006data}. We are testing how well the LLM does over many different members of ${\bm R}$ but are assuming that
other factors  we might vary make no difference. However, there are many other populations of lists that we might try, and there are many other wordings of the prompt that could be used. If we use observed success with the above prompt
to conclude that our LLM can count elements of a length-20 list with a particular success rate we are implicitly assuming that these other possible choices would make no difference. For example, an alternative to the prompt above might be:
\begin{verbatim}
prompt = ``Here is a list: '' + str(r) + ``. How many times does `mango' appear on it?''
\end{verbatim}
This would appear to be an equivalent evaluation of the task, or a modification that should make no difference. Unfortunately, this is not the case. 

As we show in Section \ref{sec:count} these assumptions most definitely do not hold. Wording of the prompt and choice of the particular items to be counted can make a substantial difference to the answer (see Table \ref{tab:count}). 
For example,  the hypothesis that tests using the two prompts given above (with everything else held constant) produce results drawn from the same distribution, is robustly rejected by a $\chi^2$ test. Thus, if we report
that our LLM can count with a particular success rate we are committing the same fixed-effect error as experimenters A and B above. 

%\subsection{Experiment, Replication and Generalization}

When we encounter a particular experimental result (e.g., $q = 0.86$ ($86.0$\%) on the $N=10$ counting task in Table \ref{tab:count}) we generally understand that this involves some margin of error. For example, rather than $q$, we expect  a repeat of the experiment to produce an estimate
$q \pm \delta_q.$
A very familiar case exploits the fact that 95\% of the values of a normal distribution lie within $1.96$ standard deviations of the mean, so we can write 
$\delta_q = 1.96 \cdot \sqrt{q\cdot (1-q) /N}$ and be confident that $95$\% of trials 
will fall in this interval \citep{taylor1982introduction}. 

However, it is important to keep track of the baked-in assumptions: this estimate  assumes that variance from sampling the list population (i.e., sampling ${\bm R}$) is the only source of randomness. If significant other sources of randomness exist, then we know only
that $\delta_q$ is greater than (and possibly much greater than) $1.96 \cdot \sqrt{q\cdot (1-q) /N}.$ That is, we have only a lower bound on our margin-of-error.
%[I don't understand the previous sentence. Need some explanation. -Shuo. Added some wording, does this help? Cormac] 
We can't rule out, for example, that the $95$\% confidence interval is $\pm 30$\%.
The results of Section \ref{sec:tasks} show that other sources of randomness that are too large to ignore do exist for several of the tasks we consider (and in some cases
appear far greater than the variance due to sampling).

%% file: listQueries/queries.tex
\section{Tasks}
\label{sec:tasks}

\subsection{Experimental setup}

In order to test LLM performance we choose tasks that have deterministic answers, and where 
 it is relatively easy to decide if the LLM gives the correct answer.  
This obviates the need for subjective assessments, heuristics, hand-labelling or error-prone parsing of the response, and allows us to scale-up testing.
The tasks we examine are: counting, finding the maximum, median and sorted version of a list of numbers, and long multiplication. The difficulty with counting and long multiplication has been observed by others \citep{arkoudas2023gpt4}.

% \begin{itemize}
% \item Counting 
% \item Finding the maximum, median and sorted version of a list of numbers
% \item Long multiplication
% \item Basic composite tasks combining the above tasks.
% \end{itemize}

Unless otherwise specified all of the conditions were evaluated on $500$ independent runs. 
Thus, for example, if a table entry reports a success rate of  89.0\% on a task, and sampling were the only source of randomness, then a reasonable estimate of the 95\% confidence interval would be $1.96 \cdot \sqrt{0.89 \times 0.11/500} \approx 2.74 \%.$ However, an important finding, below,
is that there are significant other sources of randomness, and the conventional way of estimating margins-of-error
cannot be applied.
Lists were generated independently for each trial at query-time; thus, we did not re-use lists across conditions.
All of the trials are performed using the OpenAI GPT-4 API with a temperature setting of $0.7.$
%(results for GPT-3.5 are given in Section \ref{sec:gpt35}). 
The results of all queries are available in the GitHub repository \url{https://github.com/demarinaGit/canWeCountOnLLMs}. All trials were performed using a temperature setting of 0.9.

%\subsection{Queries}

For all of the tasks we give an example prompt together with the correct answer and GPT-4's answer. Due to space constraints we show only examples where the GPT-4 response is incorrect. This is not reflective of its accuracy: in each case we give a table showing how accuracy evolves with problem size.
However, in giving examples where the answers are incorrect we illustrate that they are often very significantly \btr.%, a point that we return to in Section \ref{sec:observations}.

\input{listQueries/counts}

\input{listQueries/aggregates}

\input{listQueries/longMultiply}

%% file: listQueries/counts.tex
\subsection{Count}
\label{sec:count}
First we examine the capability of GPT-4 to perform basic counting tasks. 
We choose a length-{\tt rLen} list with two possible elements and ask 
GPT-4 to count the number of occurrences of the first element. An example query is  (let's call this wording \#1):

\chatPrompt{How many times does `mango' appear in this list: [mango, peach, peach, peach, mango, mango, mango, peach, peach, peach, mango, mango, mango].}{7}{`Mango' appears 6 times in this list.}

We evaluate for five different target lengths; the results are shown in the first column of Table \ref{tab:count}. In choosing modifications of this task we choose a different variations of the input list by replacing the word-pair `mango/peach'
with `airedale/aspidistra' (results in column 2). We alter the weights: i.e., have `mango' and `peach' appear with probabilities 70\% and 30\% instead of 50\% and 50\% (results in columns 3).
We also examine 
one simple rewording of the prompt (let's call this wording \#2):

\chatPrompt{Here is a  list: [mango, peach, peach, peach, mango, mango, mango, peach, peach, peach, mango, mango, mango]. How many times does `mango' appear on it?}{7}{`Mango' appears 6 times in this list.} 

This gives us a total of four conditions, all of which  involve the same basic counting task. We evaluate each condition with list lengths {\tt rLen}$=10, 15, 20, 30$ and $40,$ and we perform $500$ trials per condition. The results are shown in Table \ref{tab:count}. Thus,  the five rows and first four columns represent a total of $5 \times 4 \times 500 = 10,000$ 
queries to GPT-4.
 
\begin{table*}[h] 
\resizebox{1 \columnwidth}{!}{
\begin{tabular}{lcccc|lll}
\toprule {\tt rLen}   & \shortstack{Wording \#1\\Wts=[0.5,0.5]\\mango/peach} & \shortstack{Wording \#1\\Wts=[0.5,0.5]\\airedale/\\aspidistra}  & \shortstack{Wording \#1\\Wts=[0.7,0.3]\\mango/peach}  & \shortstack{Wording \#2\\Wts=[0.5,0.5]\\mango/peach}  &
  \shortstack{Comp. Cols(1,2)\\$(\chi^2,p)$}            &        \shortstack{Comp. Cols(1,3)\\$(\chi^2,p)$}             &         \shortstack{Comp. Cols(1,4)\\$(\chi^2,p)$} \\
% \toprule
% {} &  mango & airedale & xhsfgre & prob70 & reworded &           airedale &            xhsfgre &             prob70 &            reworded \\
\midrule
10 &  89.0\% &   91.2\% &  70.2\% &   96.6\% &          (1.12, 2.9e-01) &  {\bf (53.26, 2.92e-13)} &     {\bf (20.49, 6.e-06)} \\
15 &  61.2\% &   53.6\% &  31.8\% &   88.6\% &     {\bf (5.6, 1.8e-02)} &  {\bf (85.68, 2.11e-20)} &   {\bf (98.38, 3.45e-23)} \\
20 &  48.2\% &   29.6\% &  30.8\% &   76.2\% &  {\bf (35.61, 2.41e-09)} &  {\bf (30.95, 2.65e-08)} &   {\bf (82.18, 1.24e-19)} \\
30 &  12.4\% &    7.4\% &  19.0\% &   43.6\% &   {\bf (6.46, 1.10e-02)} &   {\bf (7.74, 5.41e-03)} &  {\bf (119.17, 9.60e-28)} \\
40 &  12.6\% &    7.6\% &  17.6\% &   21.0\% &   {\bf (6.34, 1.18e-02)} &   {\bf (4.49, 3.40e-02)} &   {\bf (12.03, 5.25e-04)}  \\
\bottomrule
\end{tabular} 
 }
\caption{\label{tab:count} Percent correct for counting  the occurrences of a length-{\tt rLen} list with two items chosen uniformly-at-random. Performance decays rapidly with list length.  
On the right-hand side of the table we present $\chi^2$ tests comparing the results
of the first condition with each of the others.
This test evaluates the null hypothesis that the answers  in the various conditions are drawn from the same distribution. 
Boldface entries are cases where $p < 0.05$ and we reject the null hypothesis.
The null hypothesis is robustly rejected for almost all lengths and conditions. E.g., 
when comparing columns 1 and 4 (i.e., simply switching between wording \#1 and wording \#2 with the `mango/peach' word-pair).  This demonstrates that simple modifications of the task (that might easily be assumed to make no difference) in fact are sources of variance beyond what can be explained by sampling effects.
}
\end{table*}

% \begin{table*}[h]
% \center{
% \begin{tabular}{ccccc|l}\hline
% $N$  & \shortstack{mango/\\peach} & \shortstack{airedale/\\aspidistra} & \shortstack{xhsfgre/\\jnosdfi} & Reworded & $\chi^2$ comparison of columns 1, 4\\ \hline
% 10  & 89.0\% & 91.2\% & 87.8\% & 96.6\% & $\chi^2= 22.147, df = 1, p =3.338e-6 $\\ \hline
% 15  & 61.2\% & 53.6\% & 54.8\% & 88.6\% & $\chi^2= 99.835, df = 1, p = 1.656e-23 $ \\ \hline
% 20 & 48.2\%  & 29.6\% & 42.6\% & 76.2\% & $\chi^2= 83.363, df = 1, p = 6.82e-20 $ \\ \hline
% 30 & 12.4\%  & 7.4\% & 21.4\% & 43.6\% & $\chi^2= 120.714, df = 1, p = 4.413e-28 $ \\ \hline
% 40 & 10.2\%  & 7.6\% & 14.2\% & 21.0\% & $\chi^2= 22.147, df = 1 , p = 2.52 e-6$ \\ \hline
% %100 & & 0\% & & \\ \hline
% \end{tabular}
% }
% \caption{\label{tab:count} Percent correct for counting  the occurrences of a length-$N$ list with two items chosen uniformly at random.
% The $\chi^2$ test evaluates the null hypothesis that the answers to the `mango/peach' task and the reworded version of the task are drawn from the same distribution. The null hypothesis is not supported
% for any of the lengths tested. This indicates that the two rewordings given can not be considered different instances of the same task.
% }
% \end{table*}

We use a $\chi^2$ test to determine if the responses to different ways of phrasing the task are drawn from the same distribution. For example, we can take 
the null hypothesis to be that some row of the  first and fourth columns of Table \ref{tab:count} represent answers drawn from the same distribution. E.g, for {\tt rLen}$=10$
there were $445/500$ and $483/500$ correct trials respectively. Using a standard 
$\chi^2$ test to compare these two distributions of correct/incorrect answers 
yields  ($\chi^2 = 20.49$, $df=1$, $p = 6.0 e-6$). The $p$-value can be taken as an estimate of the probability
of these results being observed if columns 1 and 4 of row 1 were produced by the same process; generally when $p < 0.05$
we say that the null hypothesis is rejected. Similarly for all the other rows, the hypothesis (that results of the task with different wording are drawn from the same distribution) is rejected. The degrees-of-freedom is $df=1$ for all of our tests since we are always doing pairwise comparisons on tasks on a binary outcome \citep{taylor1982introduction}.

The results of our $\chi^2$ tests are given in the right-hand side of Table \ref{tab:count}. 
The null hypothesis is robustly rejected for all lengths when comparing columns 1 and 4 (i.e., simply switching between wording \#1 and wording \#2 with the `mango/peach' word-pair). The null hypothesis is rejected for several lengths when comparing columns 1 and columns 2, 3 (i.e., simply switching the word-pair while using wording \#1).  This demonstrates that simple modifications of the task (that might easily be assumed to make no difference) in fact are sources of variance beyond what can be explained by sampling effects.

%\todo{[Do we need to show df (which I have no knowledge about)? It seems that showing p is sufficient? -Shuo]}

%\begin{table}[h]
%\center{
%\begin{tabular}{ccc}\hline
%$N$ & GPT-3.5-turbo & GPT-4 \\ \hline
%10 & 83\% & 99\% \\ \hline
%15 & 69\% & 80\% \\ \hline
%20 & & 67\% \\ \hline
%30 & & 12\% \\ \hline
%40 & & 9\% \\ \hline
%100 & & 0\% \\ \hline
%\end{tabular}
%}
%\caption{\label{tab:count} Percent correct for counting  the occurrences of a length-$N$ list with two items chosen uniformly at random.}
%\end{table}

%Arkoudas \citep{arkoudas2023gpt4} observes that GPT-4 fails at some basic counting tasks.
We note also that the GPT answers are biased toward under-counting. For example in the `mango/peach' case the mean of the correct answers for the five lengths tested (i.e., {\tt rLen}$=10, 15, 20, 30$ and $40$)  were:
$(5.57, 7.96, 10.57, 15.46, 20.6)$ and the GPT-4 answers were $(5.45, 7.57,  10.04, 14.09, 18.5).$ Thus, across $500$ trials, the mean GPT-4 answers were 
always lower. Among the 500 trials the ratio of over-counts:under-counts was
$(55:0, 192:2, 248:11, 428:10, 451:11).$

In the appendix  we give results using GPT-3.5, Mistral7B, and Llama7B. Note that the same basic pattern holds: i.e., the different conditions

% \subsection{Count Distinct}
% Next  we count the number of distinct occurrences of an element in a length-$N$ list with $Q=10$ items chosen uniformly at random.
% An example query is:

% \chatPrompt{How many distinct elements are in this list: [grape, apple, grape, apple, apple, plum, pear, kiwi, melon, plum, apple, plum].}{6}{There are 5 distinct elements in this list: [grape, apple, plum, pear, kiwi, melon].}

% \begin{table}[h]
% \center{
% \begin{tabular}{ccc|l}\hline
% $N$ & GPT-3.5-turbo & GPT-4 & $(\chi^2, p)$\\ \hline
% %10 & 70.1 $\pm$ 4.01\% & 75\% \\ \hline
% %15 & 44.0 $\pm$ 4.3\% & 25\% \\ \hline
% %20 & 40.1 $\pm$ 4.3\% & 5\% \\ \hline
% 10 & 70.1\% & 55.4\% & {\bf (23.111 , 1.520 e-6)}\\ \hline
% 15 & 44.0\% & 23.8\%  & {\bf (45.524, 1.507 e-11)} \\ \hline
% 20 & 40.1\% & 6.0\%  & {\bf (163.896, 1.594 e-37)}\\ \hline
% \end{tabular}
% }
% \caption{\label{tab:countDistinct} Percent correct for counting  the distinct items in a length-$N$ list with $Q$ possible items chosen uniformly at random. Note that GPT-3.5 actually does significantly better than GPT-4 on this task.}
% \end{table}

% This task represents an instance where the GPT-4 answers are significantly lower than those for GPT-3.5. 
% %\subsection{Count with non-standard separator}
% %
% %Replacing comma with `foo' and the results get worse. 

%% file: listQueries/aggregates.tex
\subsection{Maximum, Median and Sort}

Here we ask GPT-4 to perform elementary tasks on lists of numbers: return the maximum, median and sorted version of the list. 
We evaluate three different conditions. First we ask for the maximum (or median or sorted version)
of a list of {\tt rLen} numbers drawn uniformly-at-random from the interval $(100.0, 20000.00)$ and rounded to two decimals places. 
An example of the prompt for the median-finding task is:

\chatPrompt{What is the median value in this list: [7176.36, 5222.86, 1089.62, 19927.36, 5655.72, 18355.58, 18978.7, 7028.49, 14190.57, 14243.69, 11251.69]. Please write 'Answer='}{11251.69}{7176.36}

% \chatPrompt{What is the maximum value in this list: [10302, 15544, 16436, 3547, 2021, 1671, 6715, 15374, 9250, 10647, 3608, 16925, 11050, 6724, 13329, 7562, 5331, 14893, 12408, 4362, 18371, 5403, 10105, 3598, 17799, 10303, 1950, 8808, 2384, 16277, 15518, 14619, 14794, 17489, 8505, 7883, 13310, 1447, 13541, 14155, 12676, 5073, 19026, 15133, 14434, 2538, 4627, 11625, 1108, 5257].}{19026}{The maximum value in this list is 18371.}

Second,   we repeat with integers drawn uniformly-at-random from $(10, 99)$ (i.e., all list-members are 2-digit numbers). Finally, we use a list of {\tt rLen}
name-value pairs, where a randomly-chosen name is associated with a number
drawn uniformly-at-random from the interval $(100.0, 20,000.00)$ and rounded to two decimals places. An example of the latter query is:

\chatPrompt{Please sort this list in ascending order: [John: \$12158.21, Mary: \$1416.51, Peter: \$7507.58, Vivek: \$10941.54, Xian: \$10530.84, Alex: \$1641.14, Maria: \$1025.49, Frank: \$260.85, Luis: \$7464.35, Manuel: \$1782.86, Kristen: \$10085.24].}{[Frank: \$260.85, Maria: \$1025.49, Mary: \$1416.51, Alex: \$1641.14, Manuel: \$1782.86, Luis: \$7464.35,Peter: \$7507.58,  Kristen: \$10085.24, Xian: \$10530.84, Vivek: \$10941.54, John: \$12158.21]}{[Frank: \$260.85, Maria: \$1025.49, Mary: \$1416.51, Alex: \$1641.14, Manuel: \$1782.86, Peter: \$7507.58, Luis: \$7464.35, Kristen: \$10085.24, Xian: \$10530.84, Vivek: \$10941.54, John: \$12158.21]}

% \chatPrompt{Here is a list of names with their Q-value. Can you sort by ascending Q-value: [John: \$6301, Mary: \$8797, Peter: \$5832, Vivek: \$3679, Xian: \$19221, Alex: \$9459, Maria: \$7628, Frank: \$5806, Luis: \$19899, Martin: \$2157, Antonio: \$2794, Kannan: \$18793, Sasha: \$4023].}{ [Martin: \$2157, Antonio: \$2794, Vivek: \$3679, Sasha: \$4023, Frank: \$5806, Peter: \$5832, John: \$6301, Maria: \$7628, Mary: \$8797, Alex: \$9459, Kannan: \$18793, Xian: \$19221, Luis: \$19899]}
% {Here is the list of names sorted by their Q-value in ascending order: [Martin: \$2157, Antonio: \$2794, Vivek: \$3679, Sasha: \$4023, Frank: \$5806, Peter: \$5832, John: \$6301, Maria: \$7628, Alex: \$9459, Mary: \$8797, Kannan: \$18793, Luis: \$19899, Xian: \$19221].}

% \chatPrompt{What is the median value in this list: [1.67, 5.7, 5.79, 3.9, 6.71, 7.38, 5.95, 5.29, 3.48, 2.15, 1.72].}{5.29}{The median value in this list is 5.7.}

The results of the maximum, median and sorting tasks are given in Tables \ref{tab:maxTable}, \ref{tab:medianTable} and \ref{tab:sortTable} respectively. The three different list conditions are explored in columns 1-3 of these tables. As in Section \ref{sec:count}, we use a $\chi^2$ test to explore whether these different  variations on the task produce answers 
that appear drawn from the same distribution.  The right-hand portion of Tables \ref{tab:maxTable}, \ref{tab:medianTable} and \ref{tab:sortTable} gives the results; we do $\chi^2$ tests to compare columns 2 and 3  with column 1. 

Table \ref{tab:maxTable} shows the results of the maximum-finding task. Performance in all conditions is good, though not perfect
(e.g, results are almost always $> 90.0$\%). The $\chi^2$ tests show that the hypothesis that performance on the name-value version of the list is consistent with performance on the value-only list is rejected for lengths $>11$. 
The hypothesis that performance on the integer version of the list is consistent with performance on the 2-decimal floats list is rejected for all lengths.

Table \ref{tab:medianTable} shows the results of the median-finding task. Performance in all conditions is  poor
(e.g, results  are $< 90.0$\%). The $\chi^2$ tests show that the hypothesis that performance 
when the numbers are drawn from $(10.0, 20000.0)$ is consistent with performance when numbers are drawn as integers from $(10, 99)$
is rejected for all lengths.  The hypothesis that that 
name-value version of the list is consistent with performance on the value-only list is also rejected for all lengths. 
Note that the $p$-values in both cases are $\ll 0.05$, so the probability that the same process  accounts for both conditions is
very low.

Table \ref{tab:sortTable} shows the results of the sorting task. Performance in condition 2 is good, but is very poor
in condition 3 (e.g, results in  column 3 are $< 55.0$\%). The $\chi^2$ tests show that the hypothesis that performance 
when the numbers are drawn from $(10.0, 20000.0)$ is consistent with performance when numbers are drawn from $(10, 99)$
is rejected for all lengths.  The hypothesis that that 
name-value version of the list is consistent with performance on the value-only list is also rejected for all lengths. 
Again, the $p$-values indicate robust rejection of these hypotheses.

% \begin{table}[h]
% \center{
% \begin{tabular}{cccc}\hline
%  & Max. & Median & Sort \\ \hline
% 11   & 98\% & 68.4\% & 98\%\\ \hline
% 15  &  97\% & 54\% & 91\%\\ \hline
% 21  & 97\% & 35.9\% & 89\%\\ \hline 
% %31  & 91\% & 38\% & 96\%\\ \hline
% %51 & & 25\% &  \\ \hline
% %101 & & & 0\% \\ \hline 
% \end{tabular}
% }
% \caption{\label{tab:aggreg20k} Percent correct for Max, Median and Sort of a length-$N$ list of numbers. Numbers were uniformly distributed on $(100.0, 20000.0).$}
% \end{table}

% \begin{table}[h]
% \center{
% \begin{tabular}{cccc}\hline
%   & Max. & Median & Sort \\ \hline
% 11   & 99\% & 70\% & 97\%\\ \hline
% 15  & 98\%  & 48\% & 96\% \\ \hline
% 21   & 98\% & 41\% & 92\%\\ \hline 
% %31   & & \\ \hline
% %51 & & 25\% &  \\ \hline
% %101 & & & 0\% \\ \hline 
% \end{tabular}
% }
% \caption{\label{tab:aggreg20k_12sigFigs} Percent correct for Max, Median and Sort of a length-$N$ list of numbers, with 12 significant digits. Numbers were uniformly distributed on $(100.0, 20000.0).$}
% \end{table}

\begin{table*}
\center{}
 \begin{tabular}{lccc|cc}  \toprule{} 
 {\tt rLen} & \shortstack{{\tt float} in \\$(100.0,20000.0)$} & \shortstack{{\tt int} in \\$(10,99)$} & \shortstack{Name-value\\{\tt float} in \\$(100.0,20000.0)$} &        \shortstack{Compare Cols(1,2)\\$(\chi^2,p)$}            &        \shortstack{Compare Cols(1,3)\\$(\chi^2,p)$}                               \\ 
 \midrule
11 &  97.79\% &  100.0\% &  97.2\% &   {\bf (9.23, 2.38e-03)} &        (0.16, 6.93e-01) \\
15 &   96.4\% &  100.0\% &  92.2\% &  {\bf (16.35, 5.27e-05)} &  {\bf (7.44, 6.37e-03)} \\
21 &   94.4\% &  100.0\% &  86.6\% &  {\bf (26.79, 2.27e-07)} &  {\bf (16.8, 4.16e-05)} \\
\bottomrule
\end{tabular} 
%  \midrule 
% 11 &    98.0\% &    99.0\% &    99.9\% &    99.8\% &  (1.69, 1.93e-01) &   {\bf (8.69, 3.21e-03)} &   {\bf (7.45, 6.36e-03)} \\15 &    97.0\% &    98.0\% &    99.9\% &    94.0\% &  (1.03, 3.11e-01) &  {\bf (13.78, 2.06e-04)} &   {\bf (5.24, 2.21e-02)} \\
% 21 &    97.0\% &    98.0\% &    97.0\% &    86.0\% &  (1.03, 3.11e-01) &      (0.0, 1.) &  {\bf (38.89, 4.47e-10)} \\ \bottomrule\end{tabular}
 \caption{\label{tab:maxTable} Comparison of the find-maximum task. The prompt simply asks GPT-4 to find the maximum
 of a list of numbers.
 Column 1: numbers uniform on $(100.0,20000.0)$ to 2 decimals, 
  Column 2: numbers uniform on $(10,99)$ as integers,
  Column 3: name-value pairs with values uniform on $(100.0,20000.0)$ to 2 decimals.
The right-hand side of the table shows  $\chi^2$ tests comparing Column 1 to each of the others.
Boldface entries are cases where $p < 0.05$ and we reject the null hypothesis (that results in the given columns
are produced by the same process). The null hypothesis is rejected except for length-11 when comparing columns \#1 and \#3: thus
simply switching the list from numbers to name-value pairs introduces variance beyond what can be explained by sampling effects.}
\end{table*}

\begin{table*} \center{}
\begin{tabular}{lccc|ll}
 \toprule{} {\tt rLen} & \shortstack{{\tt float} in \\$(100.0,20000.0)$} & \shortstack{{\tt int} in \\$(10,99)$} & \shortstack{Name-value\\{\tt float} in \\$(100.0,20000.0)$} &        \shortstack{Compare Cols(1,2)\\$(\chi^2,p)$}            &        \shortstack{Compare Cols(1,3)\\$(\chi^2,p)$}                               \\
\midrule
11 &   68.4\% &   85.0\% &  89.6\% &  {\bf (37.62, 8.57e-10)} &   {\bf (66.46, 3.58e-16)} \\
15 &   52.8\% &   74.0\% &  89.6\% &  {\bf (47.51, 5.47e-12)} &  {\bf (163.32, 2.13e-37)} \\
21 &  35.87\% &  62.73\% &  65.6\% &  {\bf (65.82, 4.94e-16)} &   {\bf (87.12, 1.02e-20)} \\
\bottomrule
\end{tabular} 
% \midrule 
% 11 &    68.4\% &    70.0\% &    89.6\% &    84.8\% &   (0.3, 5.84e-01) &  {\bf (67.73, 1.88e-16)} &  {\bf (37.51, 9.08e-10)} \\
% 15 &    54.0\% &    48.0\% &    80.0\% &    74.0\% &   (3.6, 5.77e-02) &  {\bf (76.44, 2.27e-18)} &   {\bf (43.4, 4.46e-11)} \\
% 21 &    35.9\% &    41.0\% &    65.6\% &    62.7\% &  (2.75, 9.74e-02) &  {\bf (88.23, 5.83e-21)} &  {\bf (71.84, 2.34e-17)} \\ \bottomrule\end{tabular} 
 \caption{\label{tab:medianTable} Comparison of the find-median task. The prompt simply asks GPT-4 to find the median
 of a list of numbers.
 Column 1: numbers uniform on $(100.0,20000.0)$ to 2 decimals, 
  Column 2: numbers uniform on $(10,99)$ as integers,
  Column 3: name-value pairs with values uniform on $(100.0,20000.0)$ to 2 decimals.
The right-hand side of the table shows  $\chi^2$ tests comparing Column 1 to each of the others.
Boldface entries are cases where $p < 0.05$ and we reject the null hypothesis (that results in the given columns
are produced by the same process). The null hypothesis  for all lengths and conditions: thus
simply changing the range on the numbers, or switching to name-value pairs introduces variance beyond what can be explained by sampling effects.}
\end{table*}

%% file: listQueries/longMultiply.tex
\subsection{Long Multiply}

Here we evaluate performance at long multiplication, where we prompt the  LLM to calculate the product
of a $k_1$-digit by a $k_2$-digit number. An example for $4 \times 4$ is:

\chatPrompt{What is the product of 6438 and 9038? Please write `Answer ='}{58186644}{Answer = 58169844.}

Table \ref{tab:multiply} shows the performance multiplying a $k_1$-digit by a $k_2$-digit number for $k_1, k_2 \in \{2, 3, 4, 5\}$. Apart from the $2 \times 2$ case the results are largely poor.
Observe that perfect performance on the $2 \times 2$ task drops to negligibly correct answers for $4 \times 4.$ 

Since there is sometimes a significant difference between the $k_1 \times k_2$ result with the $k_2 \times k_1$ result
we perform a $\chi^2$ test on several of the off-diagonal elements.  The results are shown in Table \ref{tab:multiplyChi}. Note that results for the $4 \times 2$ and $2 \times 4$ are significantly different, as are those for 
$5 \times 2$ and $2 \times 5.$ Thus, even the hypothesis that performance on the $k_1 \times k_2$ multiplication 
will be equivalent to the $k_2 \times k_1$ is rejected for at least some lengths.

Both Dziri et al \citep{dziri2023faith} and Arkoudas \citep{arkoudas2023gpt4} look at the example of long multiplication. Dziri et al note that while the 
answers for $4 \times 4$ are almost always incorrect, the first and last two digits of the GPT-4 answers are almost always correct. They describe 
this as a matching of ``surface probabilities.'' That is, the first two digits of a product are determined by the leading digits of the multiplicands irrespective of length. Thus, this portion of the answer
can always be determined without paying attention to the rest. Similarly for the last few digits.

\begin{table*} 
\center{}
 \begin{tabular}{lccc|cc}\toprule{} 
 {\tt rLen} & \shortstack{{\tt float} in \\$(100.0,20000.0)$} & \shortstack{{\tt int} in \\$(10,99)$} & \shortstack{Name-value\\{\tt float} in \\$(100.0,20000.0)$} &        \shortstack{Compare Cols(1,2)\\$(\chi^2,p)$}            &        \shortstack{Compare Cols(1,3)\\$(\chi^2,p)$}                               \\
\midrule
11 &  94.93\% &  99.77\% &  52.0\% &  {\bf (18.39, 1.80e-05)} &  {\bf (231.64, 2.62e-52)} \\
15 &  94.75\% &  100.0\% &  36.0\% &  {\bf (24.48, 7.50e-07)} &  {\bf (375.91, 9.69e-84)} \\
21 &  88.32\% &   99.8\% &  15.0\% &  {\bf (56.26, 6.34e-14)} &  {\bf (528.43, 6.2e-117)} \\
\bottomrule
\end{tabular} 
 % 11 &    98.0\% &    97.0\% &    99.7\% &    61.0\% &   (1.026, 3.11e-01) &   {\bf (6.356, 1.17e-02)} &  {\bf (210.002, 1.37e-47)} \\15 &    91.0\% &    96.0\% &    98.0\% &    36.0\% &  {\bf (10.284, 1.34e-03)} &  {\bf (23.569, 1.21e-06)} &  {\bf (326.286, 6.19e-73)} \\
 % 21 &    89.0\% &    92.0\% &    99.7\% &    15.0\% &   (2.617, 1.06e-01) &  {\bf (53.693, 2.34e-13)} &  {\bf (548.478, 2.7e-121)} \\ \bottomrule\end{tabular}
 \caption{\label{tab:sortTable} Comparison of the list-sorting task. The prompt simply asks GPT-4 to sort
 a list of numbers in ascending order.
 Column 1: numbers uniform on $(100.0,20000.0)$ to 2 decimals, 
  Column 2: numbers uniform on $(10,99)$ as integers,
  Column 3: name-value pairs with values uniform on $(100.0,20000.0)$ to 2 decimals.
The right-hand side of the table shows goodness-of-fit $\chi^2$ tests comparing Column 1 to each of the others.
Boldface entries are cases where $p < 0.05$ and we reject the null hypothesis (that results in the given columns
are produced by the same process). The hypothesis that Column 2 or 3 is produced by the same process as Column 1 is rejected for all lengths: thus
simply changing the range on the numbers, or switching to name-value pairs introduces variance beyond what can be explained by sampling effects.}
\end{table*}

\begin{table}[h]
\center{
\begin{tabular}{ccccc}\hline
\diaghead(-3,2){.1em}{$k_1$}{$k_2$} & 2 & 3 & 4 &5  \\ \hline
2 & 100\% & 90.6\% & 69\% & 40.6\%\\ \hline % & 45\%\\ \hline
3 & 91.6\% & 55.2\% & 15.0\% & 6.2\% \\ \hline
4 & 80.0\% & 19.4\% & 3.2\%  & 1.0\%\\ \hline 
5 & 48.4\% & 8.2\% & 2.0\% & 0.0\%\\ \hline 
\end{tabular}
}
\caption{\label{tab:multiply} Percent correct for multiplying a $k_1$-digit by $k_2$-digit number.}
\end{table}

%Note that there appears to be a statistically significant asymmetry in the GPT-3.5-turbo results: e.g., results for a $3 \times 4$ multiplication are better than for a $4 \times 3.$

\begin{table}[h]
\center{
\begin{tabular}{cc|l}\hline
$k_1 \times k_2$ & $k_2 \times k_1$ & $(\chi^2,p)$     \\ \hline 
$3 \times 2$ & $2 \times 3$ &   $(0.308, 0.578)$   \\ \hline
$4 \times 2$ & $2 \times 4$ &   {\bf (14.863, 1.15 e-4)}   \\ \hline
$4 \times 3$ & $3 \times 4$ &   $(3.398, 0.065)$   \\ \hline
$5 \times 2$ & $2 \times 5$ &   {\bf (6.158 , 0.0130)}   \\ \hline
$5 \times 3$ & $3 \times 5$ &   $(1.496 , 0.221 )$   \\ \hline
\end{tabular}
}
\caption{\label{tab:multiplyChi} $\chi^2$ goodness-of-fit test comparing the results of a $k_1 \times k_2$ with a $k_2 \times k_1$ multiplication (i.e., the off-diagonal elements of Table \ref{tab:multiply}). }
\end{table}

%% file: listQueries/related.tex
\section{Related Work}
\label{sec:related}

It is well understood that the form of a prompt can greatly
affect the results from a LLM as a ``few-shot learner''~\citep{fewShot20},
thus giving rise to the newly minted discipline of {\em prompt engineering}.
For example, \citep{YuQS23} show that small differences in
prompting for legal reasoning tasks has a significant impact on 
the accuracy of responses.  Our results confirm these observations
for a set of simple deterministic tasks but with high statistical significance. 

On the output side, Bender et al.~\citep{bender2021dangers} note the dangers 
inherent in ascribing intent and meaning to utterances generated by LLMs.
In particular, we (as humans) make many assumptions about communications
with other humans that can easily lead us to fall prey to the 
fixed-effect fallacy when working with LLMs, potentially 
ascribing a more general capability to the LLM than
actually exists. We show that even for simple tasks there are
major sources of variance that are not easy to account for when
working with LLMs.

Our experiments with deterministic algorithms are related to 
work that examines the capability of LLMs to perform deductive reasoning
~\citep{arkoudas2023gpt4}. In these problems, as with most of the problems
we consider, the LLM must attend to most every token in the input and
not ``hallucinate'' new values that would lead to short-cut solutions
to related but different problems than the one given. In contrast to
our experiments, Arkoudas engages in a conversation with the LLM about
each of the deductive problems he poses, where the LLM often proceeds
to contradict itself upon getting a wrong answer. Indeed, the ad-hoc 
reporting of conversations with an LLM is fairly widespread~\citep{bubeck2023sparks} 
but does not rise to the level of a controlled experiment where one can 
make statistically significant statements.  Of course, for many complex
tasks it may be difficult to perform the deeper analysis we performed
here for simpler tasks.

Others have observed that LLM performance degrades when the input to the LLM
grows in size (within the limits of the LLM's context window), as we have
shown here. Interestingly, Liu et al \citep{liu2023lost} 
find that information that is 
at the beginning or end of the context window has more influence on LLM performance, 
even for simple queries that ask the LLM a question whose answer is somewhere in the
input.  That is, the position of information is another source of variance, as we
saw in the simple prompt rewording of Table~\ref{tab:count}, where the major change
was to swap the position of the input list and query (wordings \#1 and \#2). 

% Lucas Memmert, Izabel Cvetkovic, Eva A. C. Bittner:
% The More Is Not the Merrier: Effects of Prompt Engineering on the Quality of Ideas Generated By GPT-3. HICSS 2024: 7520-7529

Wu et al demonstrate considerable performance sensitivity for a series of tasks \citep{wu2023reasoning}. 
In exploring counter-factual tasks they conclude that LLMs ``rely on narrow, non-transferable procedures for
task-solving.''
Dziri et al explore failures of LLMs on seemingly trivial tasks \citep{dziri2023faith}. They are especially interested in compositional tasks. They suggest that transformers often fail since they exploit linearized patch matching rather than any 
multi-step reasoning, and that errors propagate in a fashion that compounds. 
Schaeffer et al suggest that the often-discussed emergent properties of LLMs are an artifact of the metrics chosen
rather than any fundamental improvement \citep{schaeffer2023emergent}: ``For a fixed task and a fixed model family, the researcher can choose a metric to create an emergent ability or choose a metric to ablate an emergent ability.''

Chain-of-Thought (CoT) is a prompting strategy that asks the LLM to output intermediate reasoning steps before giving the final answer. Research has found that it often improves LLM performance on complex tasks \citep{wei2022chain}. It is worth further research to understand whether CoT-style prompts are more resilient to the variations shown in our study. 

While the sensitivity of performance to prompt-phrasing has spawned the field of `prompt engineering' 
efforts to quantify this sensitivity are nascent. Sclar
et al examine the effect of phrasing on accuracy for multiple choice tasks using the LLaMA-2-13B model
\citep{sclar2023quantifying}. 
Sun et al examine zero-shot robustness on two large standardized datasets 
\citep{sun2023evaluating}. Our work extends that direction by  showing sensitivity not merely to phrasing, but also input parameter, and using GPT-4 (i.e., a far larger model than used in \citep{sclar2023quantifying,sun2023evaluating}). In focusing on tasks with arbitrarily large parameter
spaces (e.g., counting objects in lists) we avoid many of the concerns that some variant of a task
has been seen in training.

% \todo{What do we plan to do with this note? "Why is it as good as it is? Eg, length-20 lists of mango/peach probably aren't in the training corpus. Why would performance on lion/tiger be at all comparable to mango/peach?"}

Standardized exams are often used to demonstrate LLM's capabilities. For example, studies has shown GPT-4 achieving the passing criteria of the Japanese Medical Licensing Examination (JMLE) \citep{takagi2023performance}, the Uniform Bar Examination (UBE) \citep{katz2023gpt}, and the US Medical Licensing Examination (USMLE) \citep{nori2023capabilities}. Knowing that even basic tasks are sensitive to trivial variations, it is legitimate to question whether the variations between a new version of an exam and its previous versions primarily focus on factors sensitive for humans, but neglect others that can be sensitive only for LLMs.

%Graded across the UBE components, in the manner in which a human tast-taker would be, GPT-4 scores approximately 297 points, significantly in excess of the passing threshold for all UBE jurisdictions \citep{katz2023gpt}.

% Our results show that GPT-4, without any specialized prompt crafting, exceeds the passing score on USMLE by over 20 points. \citep{nori2023capabilities}.

Yarkoni \citep{yarkoni2019generalizability} argues that the problem of improper generalization goes far beyond the language issue. He suggests that 
confusing fixed effects for random ones is the source of many of 
the replication failures in the social sciences. 

Elazar et al explore the consistency of responses under rephrasing of various LLMs
\citep{DBLP:journals/corr/abs-2102-01017}. They explore general knowledge and factual questions rather than the arithmetic tasks we explore. Their findings, that all of the LLMs studied have poor consistency, are largely corroborated by our work.

Lu et al study the effect of ordering on the performance of few-shot prompts \citep{DBLP:journals/corr/abs-2104-08786}. They find that permuting the order in which examples in a few-shot prompt are presented   can make the difference between state-of-the-art and radnom performance.

%% file: listQueries/discuss.tex
\section{Discussion} 

We've shown in Section \ref{sec:tasks}, the risk that measured performance with a specific prompt fails to generalize to equivalent versions of the task. This work complements others that have documented the brittleness of GPT-4's performance (see related work in Section \ref{sec:related}). However, as far as we know, ours is the first to explore tasks with several different conditions and sufficient statistical power to rule out sampling noise as the source of observed variation. This allows us to state with some confidence that minor modifications have potentially enormous effects on measured capabilities. This problem is entirely orthogonal to the frequently mentioned difficulty with hallucinations.

Every measurement experiment comes with decisions about which factors might affect the output, and which should make no difference. Many of these decisions are implicit, and informed by our intuition and experience of the world. Since LLMs emulate  many human capabilities it is tempting to use intuitions about humans to guide decisions about
which factors should make no difference to LLM measurements. A key finding of this paper is that this assumption leads to errors that can be significant enough to invalidate claims.  Bender observes that we’ve made ``machines that can mindlessly generate text, but we haven’t learned how to stop imagining the mind behind it.'' We suggest that the dangers  of anthropomorphizing LLMs includes not just over-interpreting their capabilities, but also imagining that their robustness to variation resembles that of humans.

An interesting direction for future work is whether we can derive new margin-of-error bounds. Our problem is that the presence of unexplained variance means that
 estimating  $\delta_q = 1.96 \cdot \sqrt{q \cdot (1-q)/N}$ misses an additive component of unknown magnitude.
If rewordings of a particular task can be generated automatically then estimating their variance would allow new (albeit higher) estimates of margin-of-error.
 
Since we warn of the risks
of improper generalizations we should note the limitations of our findings. Obviously, we've explored a limited set of tasks, and a limited set of modifications of those tasks.
The tasks in this paper are chosen deliberately with several criteria. First, they are deterministic tasks with easily-determined answers; this is clearly a very restricted portion of the problems to which LLMs might be applied. Second, the tasks we choose may be 
particularly difficult for transformer architectures. That is, the attention mechanism~\citep{vaswani2017attention} decides which portions of the context window are most important in predicting the next token; however, for tasks like counting, sorting, etc., all words in the target list are important. 
%These tasks have an obvious algorithmic nature. 
Third, our prompts ask the questions in a concise and direct manner, without an attempt to guide the LLM to give a Chain-of-Thought response.

\section{Conclusion}
We have demonstrated that GPT-4 performance on simple tasks shows sensitivity to trivial modifications
and that this error can be enough to invalidate claims of capabilities.
Despite the limited scope of our experiments, 
we believe our findings point to a largely-ignored source of error that potentially affects evaluation  of LLM capabilities on all  tasks.
That is, on every task we've considered we've found that trivial modifications introduce variance that invalidates the usual margin-of-error estimates. Our evidence doesn't rule out the possibility that the problem might be larger, or smaller, or negligible on some other tasks. However,  deciding that this source of error can be ignored for a given capability comes with 
a burden-of-proof, and is something that should be demonstrated empirically, rather than just assumed. 

We find that, even when modifications are trivial and make no difference to human performance on a task, we cannot  assume that the same is true of LLM performance. In the absence of evidence to the contrary, measurements of LLM task-accuracy cannot be assumed to generalize beyond the precise conditions studied.

\section{Broader impact statment}
This paper presents work whose goal is to advance the field of Machine Learning. There are many potential societal consequences of our work, none which we feel must be specifically highlighted here.

%% file: listQueries/appendix.tex
\section{Appendix}

Here we revisit accuracy measurements for the counting task studied in Section \ref{sec:count} but using the GPT-3.5, Mistral Instruct 7B Q4 and Llama 3 8B Q4 models. GPT-3.5 was accessed via the openai API. The Mistral and Llama models were run locally using versions with quantized coefficients. Each cell in each table represents $500$ trials.  

The results for these models are show in Tables \ref{tab:countGPT35}, \ref{tab:countMistral} and \ref{tab:countLlama}. Each of these  might be compared with Table \ref{tab:count}. As can be seen, the same pattern observed in Section \ref{sec:count} holds: the null hypothesis (that accuracy in the various conditions do not differ significantly) is robustly rejected in a majority 
of cases.

\label{sec:countApp}
\begin{table*}[h] 
\resizebox{1 \columnwidth}{!}{
\begin{tabular}{lcccc|lll}
\toprule {\tt rLen}   & \shortstack{Wording \#1\\Wts=[0.5,0.5]\\mango/peach} & \shortstack{Wording \#1\\Wts=[0.5,0.5]\\airedale/\\aspidistra}  & \shortstack{Wording \#1\\Wts=[0.7,0.3]\\mango/peach}  & \shortstack{Wording \#2\\Wts=[0.5,0.5]\\mango/peach}  &
  \shortstack{Comp. Cols(1,2)\\$(\chi^2,p)$}            &        \shortstack{Comp. Cols(1,3)\\$(\chi^2,p)$}             &         \shortstack{Comp. Cols(1,4)\\$(\chi^2,p)$} \\
% \toprule
% {} &  mango & airedale & xhsfgre & prob70 & reworded &           airedale &            xhsfgre &             prob70 &            reworded \\
\midrule
10 &  78.2\% &   68.8\% &  51.6\% &   89.4\% &  {\bf (10.86, 9.81e-04)} &   {\bf (76.49, 2.22e-18)} &  {\bf (22.28, 2.35e-06)} \\
15 &  61.8\% &   35.4\% &  19.6\% &   55.6\% &   {\bf (68.7, 1.15e-16)} &  {\bf (182.72, 1.23e-41)} &         (3.71, 5.40e-02) \\
20 &  17.8\% &    7.4\% &   6.8\% &   33.0\% &  {\bf (23.62, 1.17e-06)} &   {\bf (27.03, 2.00e-07)} &  {\bf (29.69, 5.08e-08)} \\
30 &  12.2\% &    5.0\% &   6.8\% &   13.8\% &  {\bf (15.58, 7.89e-05)} &    {\bf (7.86, 5.05e-03)} &         (0.43, 5.10e-01) \\
40 &   7.8\% &    2.2\% &   2.4\% &    8.8\% &  {\bf (15.35, 8.94e-05)} &   {\bf (13.97, 1.86e-04)} &         (0.21, 6.47e-01) \\
\bottomrule
\end{tabular} 
 }
\caption{\label{tab:countGPT35} GPT-3.5 Percent correct for counting  the occurrences of a length-{\tt rLen} list with two items chosen uniformly-at-random. Performance decays rapidly with list length.  
On the right-hand side of the table we present $\chi^2$ tests comparing the results
of the first condition with each of the others.
This test evaluates the null hypothesis that the answers  in the various conditions are drawn from the same distribution. 
Boldface entries are cases where $p < 0.05$ and we reject the null hypothesis.
The null hypothesis is robustly rejected for almost all lengths and conditions. 
}
\end{table*}

\begin{table*}[h] 
\resizebox{1 \columnwidth}{!}{
\begin{tabular}{lcccc|lll}
\toprule {\tt rLen}   & \shortstack{Wording \#1\\Wts=[0.5,0.5]\\mango/peach} & \shortstack{Wording \#1\\Wts=[0.5,0.5]\\airedale/\\aspidistra}  & \shortstack{Wording \#1\\Wts=[0.7,0.3]\\mango/peach}  & \shortstack{Wording \#2\\Wts=[0.5,0.5]\\mango/peach}  &
  \shortstack{Comp. Cols(1,2)\\$(\chi^2,p)$}            &        \shortstack{Comp. Cols(1,3)\\$(\chi^2,p)$}             &         \shortstack{Comp. Cols(1,4)\\$(\chi^2,p)$} \\
% \toprule
% {} &  mango & airedale & xhsfgre & prob70 & reworded &           airedale &            xhsfgre &             prob70 &            reworded \\
\midrule 
10 &  37.6\% &   59.2\% &  28.6\% &   40.8\% &  {\bf (45.84, 1.28e-11)} &  {\bf (8.74, 3.11e-03)} &         (0.94, 3.31e-01) \\
15 &  14.0\% &   30.8\% &  15.6\% &   35.6\% &  {\bf (39.63, 3.07e-10)} &        (0.39, 5.33e-01) &  {\bf (61.39, 4.68e-15)} \\
20 &  16.8\% &    6.2\% &  22.4\% &   40.2\% &  {\bf (26.57, 2.54e-07)} &  {\bf (4.63, 3.15e-02)} &  {\bf (66.03, 4.43e-16)} \\ 
\bottomrule
\end{tabular} 
 }
\caption{\label{tab:countMistral} Mistral Percent correct for counting  the occurrences of a length-{\tt rLen} list with two items chosen uniformly-at-random. Performance decays rapidly with list length.  
On the right-hand side of the table we present $\chi^2$ tests comparing the results
of the first condition with each of the others.
This test evaluates the null hypothesis that the answers  in the various conditions are drawn from the same distribution. 
Boldface entries are cases where $p < 0.05$ and we reject the null hypothesis.
The null hypothesis is robustly rejected for a majority of lengths and conditions.  
}
\end{table*}

\begin{table*}[h] 
\resizebox{1 \columnwidth}{!}{
\begin{tabular}{lcccc|lll}
\toprule {\tt rLen}   & \shortstack{Wording \#1\\Wts=[0.5,0.5]\\mango/peach} & \shortstack{Wording \#1\\Wts=[0.5,0.5]\\airedale/\\aspidistra}  & \shortstack{Wording \#1\\Wts=[0.7,0.3]\\mango/peach}  & \shortstack{Wording \#2\\Wts=[0.5,0.5]\\mango/peach}  &
  \shortstack{Comp. Cols(1,2)\\$(\chi^2,p)$}            &        \shortstack{Comp. Cols(1,3)\\$(\chi^2,p)$}             &         \shortstack{Comp. Cols(1,4)\\$(\chi^2,p)$} \\
% \toprule
% {} &  mango & airedale & xhsfgre & prob70 & reworded &           airedale &            xhsfgre &             prob70 &            reworded \\
\midrule 
10 &  37.4\% &   25.2\% &  31.2\% &   23.0\% &  {\bf (16.74, 4.28e-05)} &  {\bf (3.99, 4.57e-02)} &  {\bf (23.91, 1.01e-06)} \\
15 &  11.4\% &    4.2\% &   5.8\% &    2.6\% &  {\bf (17.03, 3.67e-05)} &  {\bf (9.27, 2.32e-03)} &   {\bf (28.4, 9.85e-08)} \\
20 &   2.2\% &    0.6\% &   0.4\% &    0.0\% &         (3.55, 5.96e-02) &  {\bf (4.99, 2.55e-02)} &   {\bf (9.19, 2.43e-03)} \\ 
\bottomrule
\end{tabular} 
 }
\caption{\label{tab:countLlama} Llama Percent correct for counting  the occurrences of a length-{\tt rLen} list with two items chosen uniformly-at-random. Performance decays rapidly with list length.  
On the right-hand side of the table we present $\chi^2$ tests comparing the results
of the first condition with each of the others.
This test evaluates the null hypothesis that the answers  in the various conditions are drawn from the same distribution. 
Boldface entries are cases where $p < 0.05$ and we reject the null hypothesis.
The null hypothesis is robustly rejected for almost all lengths and conditions.  
}
\end{table*}